# Robust Meta-Model for Predicting the Need for Blood Transfusion in Non-traumatic ICU Patients

Alireza Rafiei[a,∗], Ronald Moore[a], Tilendra Choudhary[b], Curtis Marshall[b], Geoffrey Smith[b], John D. Roback[b], Ravi M. Patel[b], Cassandra D. Josephson[b,c,d] and Rishikesan Kamaleswaran[b]

[a]*Department of Computer Science, Emory University, Atlanta, 30322, Georgia, USA*

[b]*Department of Biomedical Informatics, Emory University School of Medicine, Atlanta, 30322, Georgia, USA*

[c]*Department of Oncology, The Johns Hopkins University School of Medicine, Baltimore, 21205, Maryland, USA*

[d]*Cancer and Blood Disorders Institute, Johns Hopkins All Children's Hospital, St. Petersburg, 33701, Florida, USA*

## ARTICLE INFO

## ABSTRACT

**Objective**: Blood transfusions are crucial in the intensive care unit (ICU) to address anemia and coagulopathy. Accurately predicting blood transfusion needs is essential for optimal resource allocation and identifying critically ill patients at risk of various non-traumatic blood deficiency and post-operative events. However, existing clinical decision support systems have primarily targeted a particular patient demographic with unique medical conditions, concentrating on a single type of blood transfusion. This study aims to develop an advanced machine learning-based model to predict the probability of transfusion necessity in the next 24-hour period for all non-traumatic ICU patients with different medical conditions.
**Methods**: This retrospective cohort study utilizes pre-transfusion laboratory values and vital signs of non-traumatic ICU patients for the development of machine learning predictors. Our approach involved the analysis of an extensive dataset comprising 72,072 adult patient encounters at a high-volume metropolitan academic hospital in the USA from 2016 to 2020 for various medical reasons. We developed a meta-learner alongside diverse machine learning models to serve as predictors. These models were created on an annual basis; we trained them using data from four years and evaluated their performance on the data from the remaining, unseen year, iterating this process five times.
**Results**: The experimental results revealed that the meta-model surpasses the other models in different development scenarios. It achieved notable performance metrics, including an Area Under the Receiver Operating Characteristic (AUROC) curve of 0.97, an accuracy rate of 0.93, and an F1-score of 0.89 in the best scenario.
**Conclusion**: The development of machine learning models for the first time has been applied to predict transfusions in a diverse cohort of critically ill patients. The findings of this evaluation confirm that our model is not only able to predict the need for transfusion effectively but also provides biomarkers useful for transfusion decisions.

## 1. Introduction

Patients in the intensive care unit (ICU) frequently develop anemia or coagulopathy that is associated with adverse outcomes, such as increasing risk of life-threatening situations, thrombosis, and coronary artery diseases.[1] Post-surgical and accident-affected patients also suffer from a high risk of mortality due to severe blood loss. Transfusion of blood components is generally recommended as a clinical treatment in such scenarios. Massive blood transfusions (MTs) are essential for patients with uncontrolled intraoperative hemorrhage to avoid complications. The MT protocol (MTP) is commonly applied to trauma patients. In transfusion medicine, *Trauma* typically refers to physical injury or bleeding due to an accident or surgery. In contrast, non-traumatic blood transfusions are needed for a variety of clinical reasons that are not associated with physical injuries or trauma.

The reasons include healthy blood cell deficiency, anemia, coagulopathy, and other disorders (e.g., thrombocytopenia, hemophilia, kidney or liver disease, severe infection, and sickle cell disease). However, identification of non-traumatic ICU patients requiring transfusions is more difficult than identifying traumatic patients requiring massive transfusions. Compared to all other blood products, resuscitation with red blood cell (RBC) components is most common and frequent in transfusion patients. Approximately 85 million RBC units are transfused each year worldwide, and about 15 million are annually transfused in the United States.[2] In clinical practices, physicians often make decisions for blood transfusion primarily based on a few lab-screening features of a patient, such as anemia symptoms, hemoglobin levels, and platelet count. For example, the need for RBC transfusion is mostly decided by a hemoglobin threshold level of 7 to 8 g/dL, also suggested by the American Association of Blood Banks (AABB).[2] However, in urgent scenarios of ICU, clinicians may not be able to exhaustively evaluate all markers of a patient, such as clinical history, lab values, and demographics, which can be important. Delayed infusion, improper dosage and type of blood-products selection in transfusion may even degrade the patient's health. Thus, devising an efficient decision-making tool is critical to optimize the treatment strategies for blood transfusion of ICU patients.

Numerous research studies on predicting RBC transfusion are well-documented in the literature. The techniques used in these works vary from clinical measures[3] and standard regression analysis[4,5] to more complex machine learning methods such as neural networks[6,7,8,9] and reinforcement learning.[1] It is important to note that the majority of these prior studies were focused on the transfusion of patients undergoing specific operations, including cardiovascular surgery,[10,11,12] head and neck surgery,[13] liver transplan-tation,[14] prostatectomy,[15] and hip fracture surgery.[16] Ad-ditionally, most of the previous literature on blood trans-fusion prediction had incorporated patient demographics into model development,[8,16,9,6,17,10,18,11,5,13] which may lead to biased predictions during evaluation. Fortunately, informative routinely collected laboratory tests are avail-able to aid in the development of these models, includ-ing hemoglobin, hematocrit, platelet count, white blood cell count, creatinine, international normalized ratio (INR), bilirubin, partial thromboplastin time (PTT). However, ex-isting works use a small subset of these lab values in their predictive model developments. Therefore, it is imperative to perform a more generalized analysis for all kinds of non-bleeding ICU patients, irrespective of diagnoses and demographic variables.

In this study, a unique combination of parameterized machine learning-based schemes and significantly compre-hensive clinical features were employed to devise the de-cision model for blood transfusion requirement prediction in critical care units. To broaden the understanding of the rationale behind transfusion needs and to enhance predic-tion efficiency, we explored different parameterized machine learning-based schemes, utilizing an extensive set of clinical features, to develop a clinical support decision system for transfusion requirement prediction in critically ill patients. The research centers on pinpointing which ICU patients will most likely need a blood transfusion in the following 24 hours. For this aim, we proposed a generalizable and interpretable meta-model capable of predicting the need for transfusions of various blood products, including RBC, Plasma, and Platelets. The general workflow for our pro-posed architecture can be viewed in Figure 1.

Our contributions are as follows:

- Conduct a comprehensive analysis on a large scale of non-traumatic critically ill patient cohorts with different medical conditions over five years.

- Propose a meta-model for transfusion prediction that develops generalizable knowledge of transfusion pa-tients.

- Feature importance analysis of the meta-model to interpret reasoning behind the model's transfusion predictions.

## 2. Material and Methods

### 2.1. Data Collection

Physiological data was continuously acquired and archived using the BedMaster (Excel Medical, Jupiter, FL) software from 150 ICU beds at Emory University Hospital (Atlanta, GA). Many clinical features were collected continuously at a sampling interval of 1-hour from a given patient's admission through to discharge. However, some were derived from the electronic health records of enrolled patients. Extracted clinical features consist of vital signs and lab values from complete blood count (CBC), hepatic, pancreatic, cardiac, arterial blood gas (ABG), and inflammation tests. In this retrospective study, up to 24 hours of data preceding transfu-sion initiation was used for transfused patients admitted from 2016 to 2020, containing 72,072 patient encounters. Clinical data of the 24-hour timing window after the admission was considered for other non-transfused patients. Depending on the severity, each patient may undergo multiple transfusions, and thus, for every patient, clinical features were median-aggregated in their processing windows to have single en-tries per transfusion.

In this study, adult ICU non-trauma patients transfused with RBC, platelets, plasma, or whole blood products were included in the *Transfused* cohort. We excluded massively transfused patients showing bleeding/traumatic complica-tions by discarding those who received more than three transfusions in a continuous 6-hour window. Whereas all the adult ICU patients without any blood transfusion were included under the *Non-transfused* group. Patients with in-adequate data for processing and having all the features miss-ing were removed from the study. Finally, the study included a total of 18,314 transfused and 53,758 non-transfused en-counters. Demographic distribution and clinical statistics of involved patients are summarized in Table 1. For better gen-eralization, our study involves patients from various hospital departments and surgery sections. All transfusion and non-transfusion patients' distribution characterized by clinical features is shown by a Uniform Manifold Approximation and Projection for Dimension Reduction (UMAP) representation in Figure 2, where color labels depict various hospital ser-vice sections.

### 2.2. Data Processing

In this study, a year-wise analysis was performed for patients admitted to Emory Hospital ICU over a five-year span, from 2016 to 2020. In routinely collected lab variables and vital signs, we discarded variables missing more than 90% of values. Subsequently, a total of 43 clinical variables were selected as independent and robust features from Pearson's cross-correlation analysis. Supplemental Table 1 displays these features along with their respective units of measure-ment. The Multivariate Imputation by Chained Equations (MICE) algorithm was used to impute the missing values in features.[19] We then applied a min-max scaler to standardize the input data feature range. Subsequently, principal com-ponent analysis (PCA) was employed to reduce dimension-ality, mitigate noise, and simplify the dataset. We selected the number of principal components that together explain 90% of the variability within the original dataset. In the initial experiment, models were trained on the 2017 to 2020 datasets and then evaluated on the 2016 dataset. In order to show temporal consistency, we conducted it iteratively on an annual basis.

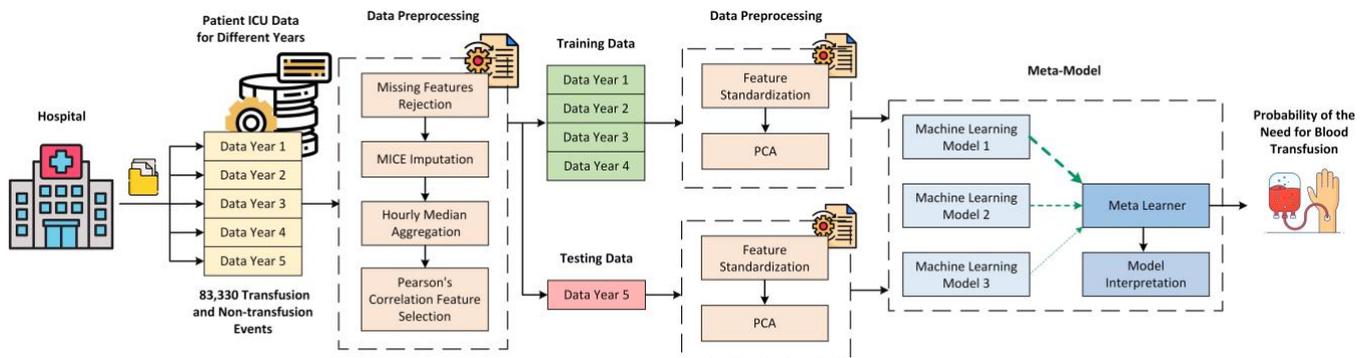

**Figure 1:** Workflow diagram of the proposed architecture. Electronic health records data collected from Emory University Hospital is preprocessed using missing features rejection, MICE imputation, aggregation, and Pearson's correlation feature selection. One year of data is used for testing, while the other years of data are used for training. The data is then further preprocessed using feature standardization and principal component analysis before being input into the meta-model for development, evaluation, and model interpretation.

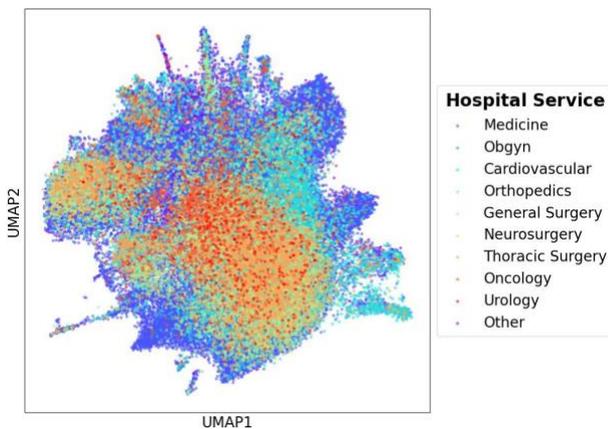

**Figure 2:** UMAP presenting all transfusion and non-transfusion events, characterized by clinical values, in 2016-2020 from various hospital services. Note that OBGYN refers to Obstetrics and Gynecology.

## 2.3. Machine Learning Models

We utilized five distinct machine learning algorithms to predict the probability of necessity for blood transfusions 24 hours in advance during ICU stays. These included logistic regression (LR), random forest (RF), feedforward neural networks (FNN), support vector machines (SVM), and XG-Boost (XGB). To improve the predictive performance of the blood transfusion need, a meta-model was constructed, forming a stacking ensemble model grounded in the principle of stacked generalization.[20,21] This technique harnesses the collective predictive strength of various models by aggregating individual predictions into a cohesive final prediction through a meta-model. This wisdom of the crowd approach aims to enhance different predictive performance metrics with the amalgamation of multiple base models. During the implementation, we tried different combinations of the developed based models and ultimately selected the RF, SVM, and XGB as the first-level models. Each model contributed its unique predictive strengths to the ensemble, with the objective of enhancing the overall accuracy of the final prediction. We also conducted a thorough examination of various meta-learners for transfusion need prediction to assess their efficacy in integrating the first-level models' predictions. LR, RF, Adaboost, voting classifier, a three-layer FNN, and Gaussian Naïve Bayes (NB) were analyzed. The Gaussian NB model was finally chosen as the meta-model.

To identify the optimal set of hyperparameters for the machine learning models, we undertook an extensive search that covered the most impactful parameters across the different models. Supplemental Table 2 details the hyperparameters and their associated values analyzed using a grid search strategy to pinpoint the optimal hyperparameters. Our primary performance metric was the area under the receiver operating characteristic curve (AUROC). AUROC can encapsulate a more holistic view of the classification performance of a model and is not biased by the imbalanced class distribution. As a result, models with a higher AUROC potentially lead to more efficient models in the prediction of blood transfusion by maintaining the balance between specificity and sensitivity metrics. Eventually, the performance of the developed models was assessed using AUROC, accuracy, F1-score, precision, and recall.

We considered five unique scenarios for training and evaluating the machine learning models on a year-by-year basis. Specifically, each model was trained using data from a four-year period and then tested on data from a subsequent, distinct hold-out year. For instance, one of the scenarios involved training the models on data collected from 2016 to 2019 and then testing them on data from 2020. All the experiments were conducted on Python 3.8.8 with scikit-learn 1.3.0, utilizing an NVIDIA GeForce GTX 950M graphics

**Table 1**
Cohort characteristics for patients admitted to the hospital from 2016 to 2020.

| Characteristic | Total encounters n = 72072 (100%) | Non-transfused n = 53758 (74.6%) | †Transfused n = 18314 (25.4%) | *p-value |
|---|---|---|---|---|
| Age, median [95% CI] | 63.0 [25.0, 90.0] | 62.0 [24.0, 90.0] | 64.0 [26.0, 88.0] | <0.001 |
| Gender, n (%) | | | | |
|     Female | 33985 (47.2) | 24834 (46.2) | 9151 (50.0) | <0.001 |
|     Male | 38087 (52.8) | 28924 (53.8) | 9163 (50.0) | |
| Race, n (%) | | | | |
|     African American or Black | 29833 (41.4) | 22107 (41.1) | 7726 (42.2) | 0.012 |
|     Caucasian or White | 36317 (50.4) | 27263 (50.7) | 9054 (49.4) | |
|     Other | 5922 (8.2) | 4388 (8.2) | 1534 (8.4) | |
| Ethnicity, n (%) | | | | |
|     Hispanic or Latino | 2226 (3.1) | 1679 (3.1) | 547 (3.0) | 0.303 |
|     Non-Hispanic or Latino | 64667 (89.7) | 48180 (89.6) | 16487 (90.0) | |
|     Other | 5179 (7.2) | 3899 (7.3) | 1280 (7.0) | |
| Hospital Service, n (%) | | | | |
|     Medicine | 32245 (44.7) | 25212 (46.9) | 7033 (38.4) | <0.001 |
|     OBGYN | 323 (0.4) | 219 (0.4) | 104 (0.6) | |
|     Cardiovascular | 13416 (18.6) | 10396 (19.3) | 3020 (16.5) | |
|     Orthopedics | 1538 (2.1) | 1088 (2.0) | 450 (2.5) | |
|     General Surgery | 2417 (3.4) | 1349 (2.5) | 1068 (5.8) | |
|     Neurosurgery | 4643 (6.4) | 4019 (7.5) | 624 (3.4) | |
|     Thoracic Surgery | 4265 (5.9) | 2693 (5.0) | 1572 (8.6) | |
|     Oncology | 1310 (1.8) | 677 (1.3) | 633 (3.5) | |
|     Urology | 363 (0.5) | 236 (0.4) | 127 (0.7) | |
|     Other | 11552 (16.0) | 7869 (14.6) | 3683 (20.1) | |
| In-Hospital Mortality, n (%) | 4888 (6.8) | 2932 (5.5) | 1956 (10.7) | <0.001 |
| Height (cm), median [95% CI] | 170.2 [149.9, 190.5] | 170.2 [149.9, 190.5] | 169.0 [149.9, 190.5] | <0.001 |
| Weight (kg), median [95% CI] | 81.0 [45.6, 145.0] | 82.0 [45.7, 147.4] | 78.3 [45.4, 136.4] | <0.001 |
| Albumin, median [95% CI] | 3.4 [2.0, 4.6] | 3.6 [2.2, 4.7] | 3.0 [1.7, 4.3] | <0.001 |
| BUN, median [95% CI] | 19.0 [6.0, 89.0] | 18.0 [6.0, 84.0] | 23.0 [6.0, 100.0] | <0.001 |
| Creatinine, median [95% CI] | 1.0 [0.5, 9.9] | 1.0 [0.5, 10.0] | 1.1 [0.4, 9.5] | <0.001 |
| Hemoglobin, median [95% CI] | 10.9 [6.6, 15.9] | 11.7 [8.0, 16.2] | 7.8 [5.5, 13.4] | <0.001 |
| Lactic Acid, median [95% CI] | 1.5 [0.6, 7.1] | 1.5 [0.6, 6.2] | 1.5 [0.6, 9.0] | <0.001 |
| Lipase, median [95% CI] | 26.0 [3.0, 465.0] | 25.0 [3.0, 505.1] | 27.0 [3.0, 390.8] | <0.001 |
| Methemoglobin, median [95% CI] | 0.4 [0.1, 1.2] | 0.3 [0.0, 1.0] | 0.5 [0.1, 1.4]] | <0.001 |
| SpO$_2$/FiO$_2$ Ratio, median [95% CI] | 250.0 [96.0, 476.2] | 250.0 [95.5, 476.2] | 247.8 [97.0, 476.2] | <0.001 |
| Platelets, median [95% CI] | 210.0 [44.0, 481.0] | 217.0 [83.0, 459.0] | 179.0 [15.0, 534.0] | <0.001 |
| PTT, median [95% CI] | 31.2 [22.3, 108.5] | 30.9 [22.3, 115.5] | 31.9 [22.3, 102.6] | <0.001 |

Abbreviations used – BUN: blood urea nitrogen, FiO$_2$: fraction of inspired oxygen, OBGYN: obstetrics and gynecology, PTT: partial prothrombin time, SpO$_2$: peripheral blood oxygen saturation, [95% CI]: 95 percent confidence interval. Note that the listed dynamic features, including lab values and vital signs, are based on pre-transfusion data for transfused patients and post-admission data for non-transfused patients.

* P-values for Gender, Race, Ethnicity, Hospital Service, and In-Hospital Mortality were computed using the Chi-square test. All other p-values were computed using the Kruskal-Wallis test.

† Transfused column has data of all patient encounters who received at least one transfusion with no MTP. However, dynamic clinical variables were presented here by considering their index transfusions only.

card, an Intel Core i7 processor at 2.60GHz, and 16GB of RAM.

## 3. Results and Discussion

### 3.1. Patient Cohort Characteristics

Table 1 contains the characteristics of the patient cohorts, particularly of ICU patients with no active bleeding who received at least one transfusion and those who did not receive any transfusion. It can be seen that there are no significant differences between the transfused and non-transfused patients for the lactic acid and most demographic variables. However, there are significant differences for the remaining variables in the table. Patients who received a transfusion had slightly higher creatinine levels, lower lipase levels, and lower SpO$_2$/FiO$_2$ ratios than their non-transfused counterparts. Additionally, those who received a transfusion also had lower hemoglobin levels and lower platelet counts than those who did not receive a transfusion. This is consistent with the transfusion criteria outlined by.[2]

**Table 2**
Performance metrics of the developed machine learning models across different model development scenarios.

| Year | 2016 | | | | | 2017 | | | | | 2018 | | | | | 2019 | | | | | 2020 | | | | |
|---|---|---|---|---|---|---|---|---|---|---|---|---|---|---|---|---|---|---|---|---|---|---|---|---|---|
| Metric | AUC | Acc | F1 | Pre | Rec | AUC | Acc | F1 | Pre | Rec | AUC | Acc | F1 | Pre | Rec | AUC | Acc | F1 | Pre | Rec | AUR | Acc | F1 | Pre | Rec |
| LR | 0.93 | 0.88 | 0.83 | 0.84 | 0.82 | 0.94 | 0.90 | 0.85 | 0.85 | 0.85 | 0.94 | 0.90 | 0.86 | 0.84 | 0.87 | 0.93 | 0.88 | 0.84 | 0.83 | 0.84 | 0.93 | 0.88 | 0.84 | 0.86 | 0.82 |
| FR | 0.94 | 0.88 | 0.83 | 0.83 | 0.83 | 0.95 | 0.90 | 0.85 | 0.85 | 0.84 | 0.95 | 0.89 | 0.85 | 0.84 | 0.86 | 0.94 | 0.88 | 0.84 | 0.83 | 0.84 | 0.95 | 0.88 | 0.83 | 0.89 | 0.80 |
| FNN | 0.93 | 0.88 | 0.82 | 0.85 | 0.78 | 0.94 | 0.89 | 0.82 | 0.88 | 0.77 | 0.94 | 0.88 | 0.83 | 0.83 | 0.83 | 0.95 | 0.89 | 0.85 | 0.87 | 0.83 | 0.92 | 0.86 | 0.81 | 0.84 | 0.78 |
| XGB | 0.95 | 0.89 | 0.84 | 0.86 | 0.82 | 0.96 | 0.91 | 0.86 | 0.87 | 0.85 | 0.95 | 0.90 | 0.86 | 0.86 | 0.87 | 0.95 | 0.89 | 0.85 | 0.88 | 0.83 | | 0.89 | 0.83 | 0.89 | 0.78 |
| SVM | 0.95 | 0.89 | 0.84 | 0.88 | 0.80 | 0.95 | 0.91 | 0.86 | 0.89 | 0.83 | 0.96 | 0.91 | 0.87 | 0.87 | 0.87 | 0.95 | 0.90 | 0.85 | 0.86 | 0.83 | 0.93 | 0.87 | 0.82 | 0.90 | 0.75 |
| MM | 0.95 | 0.89 | 0.84 | 0.85 | 0.84 | 0.96 | 0.91 | 0.86 | 0.87 | 0.86 | 0.97 | 0.93 | 0.89 | 0.90 | 0.89 | 0.95 | 0.89 | 0.85 | 0.86 | 0.84 | 0.94 | 0.88 | 0.84 | 0.88 | 0.81 |

Out of 72,072 included patient encounters between 2016 and 2020 in the study, 18,314 received transfusions, while 53,758 did not receive any. Among all years, the highest number of transfusions were noted in 2019 and 2020, the COVID-19 years, with counts of 6504 and 6515, respectively. Also, the average number of transfusions received by each transfusion encounter was 1.72 in 2019 and 1.66 in 2020. We hypothesize that COVID-19 might be the driving factor for rapid health deterioration, leading to the increased number of transfusions during these years.

Additionally, to reveal the correlation between hemoglobin levels and receiving blood transfusion, Supplemental Figure 2 presents a boxplot demonstrating the distribution of hemoglobin levels in both transfused and non-transfused cohorts. A Pearson's correlation coefficient of 0.675 was obtained (p<0.001). When considering 7 g/dL as a threshold for transfusion initiation, it is observed that patients with hemoglobin levels quite above this mark also received transfusions, and patients with hemoglobin less than this mark also did not get transfused. This highlights the insufficiency of relying solely on hemoglobin levels to develop an efficient transfusion decision support system.

## 3.2. Performance Results and Analysis

The performance results of five different test scenarios are presented in Table 2, where the specified year denotes the evaluation period. Figure 3 shows the combined receiver operating characteristic (ROC) and precision-recall curves of the developed models for all five development scenarios. Of note, we calculated and plotted the mean with the standard deviation of all five scenarios for each data point of the models.

Overall, the meta-model consistently outperformed other models across various scenarios, maintaining an AUROC of at least 0.94. It exhibited well-shaped ROC and precision-recall curves, while also other models can demonstrate comparable curve shapes. Among the rest, the SVM, XGB, and FNN models registered the best performance. Specifically, the SVM model excelled in terms of precision across different scenarios, while the meta-model had the highest recalls. When evaluated on unseen data from the year 2018 and trained on data from other years, the meta-model achieved an impressive performance, boasting an AUROC of 0.97, an accuracy rate of 0.93, and an F1 score of 0.89. Supplemental Figure 1 illustrates the calibration plot of the different developed models for various development scenarios. This plot reveals that all of the developed models are relatively well-calibrated.

Figure 4 presents the hierarchical SHapley Additive explanations (SHAP) panel of the meta-model evaluated on the 2020 data.[22,23] It offers valuable insight into how the meta-model relies on its base models to predict the necessity of a transfusion for a given patient. Notably, the prediction output from the RF algorithm stands out as the most influential model affecting the meta-model's decisions. The second column of the panel further delineates the impact of the top five features within each of the three baseline models on their final predictions. Across the board, hemoglobin and platelets emerge as the most significant features in the individual machine learning models and, subsequently, the overarching meta-model. Additionally, the SHAP scatter plots provide a visual representation of the influence exerted by different features on specific predictions, illustrating both the magnitude and direction of that influence. It should be noted that the SHAP panel for the meta-model, when evaluated across different years, exhibited largely similar patterns, with only minor variations. The 2020 scenario was visualized arbitrarily as an example.

Currently, the proposed study is limited to predicting the need for blood transfusions only. Despite this limitation, the study represents a pioneering effort in predicting the need for different types of transfusion and non-trauma ICU patients with various medical conditions. The capacity to utilize a vast array of heterogeneous training data makes the algorithms more robust in the face of incomplete, noisy ICU data, and simulating different 'use cases' to refine parameters is a crucial step in addressing the unique challenges associated with ICU research. Important next steps include extending the decision-making model's output to encompass not only an estimation of blood transfusion requirements but also the prediction of the required type of blood product. Additionally, integrating the prediction of the volume and rate of transfusion into these models could be beneficial. The next phases of this research will involve analyzing patients' longitudinal data and conducting a prospective study. This will enable the deployment of the best-performing model in real ICU settings and allow for its performance to be enhanced through iterative optimizations. A use-case scenario for deploying the proposed workflow as a clinical decision support system in the ICU settings for providing real-time predictions is shown in Figure 5.

## 4. Conclusion

In this study, we developed machine learning-based prediction models for identifying critical care patients most likely to require blood transfusion. For this aim, a unique combination of clinical features and parameterized models were explored and established. The utilization of pre-transfusion laboratory values and vital signs as features had been instrumental in the development of these models. The emphasis was placed on creating a meta-learner that was not only generalizable across different patient populations but also offered clear interpretative value in its predictions regarding transfusion necessities. Our dataset consisted of a comprehensive array of transfusion-related events from over 70,000 adult patient encounters representing a broad spectrum of medical conditions, all of whom were treated at the Emory University Hospital. However, our model needs to be cross-validated with other hospitals for more generalization. Hence, future endeavors will aim to validate extensively and integrate these models into clinical workflows and assess their effectiveness on a broader scale, with the ultimate goal of refining and personalizing care in critical settings.

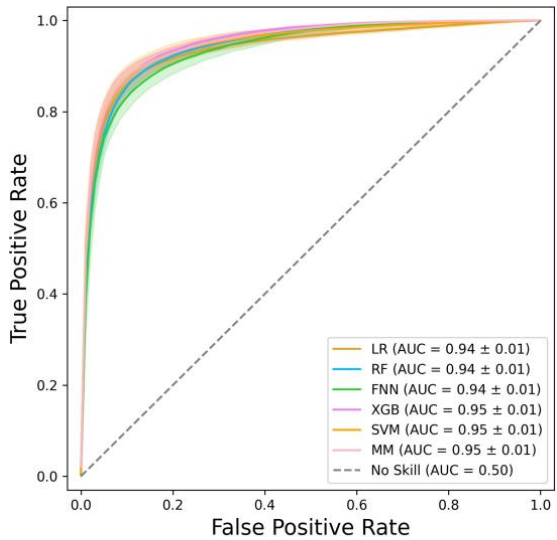

(a)

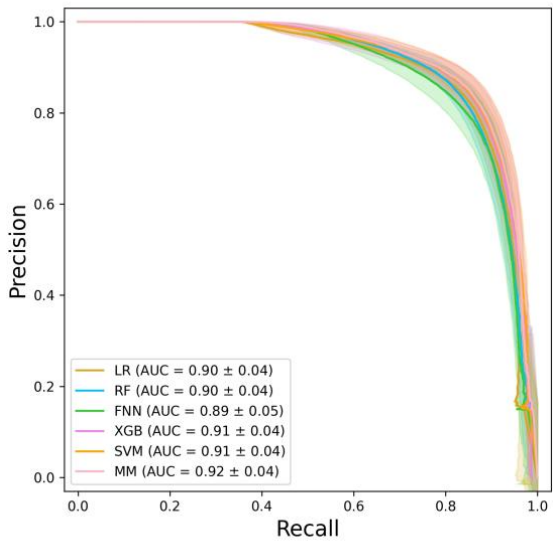

(b)

**Figure 3:** (a) ROC curves and (b) precision-recall curves of the machine learning models for transfusion need prediction in the five development scenarios. The curves are represented by a solid line indicating the mean, with the 95% confidence interval depicted as a shaded area.

## Declaration of Competing Interest

The authors declare that they have no known competing financial interests or personal relationships that could have appeared to influence the work reported in this paper.

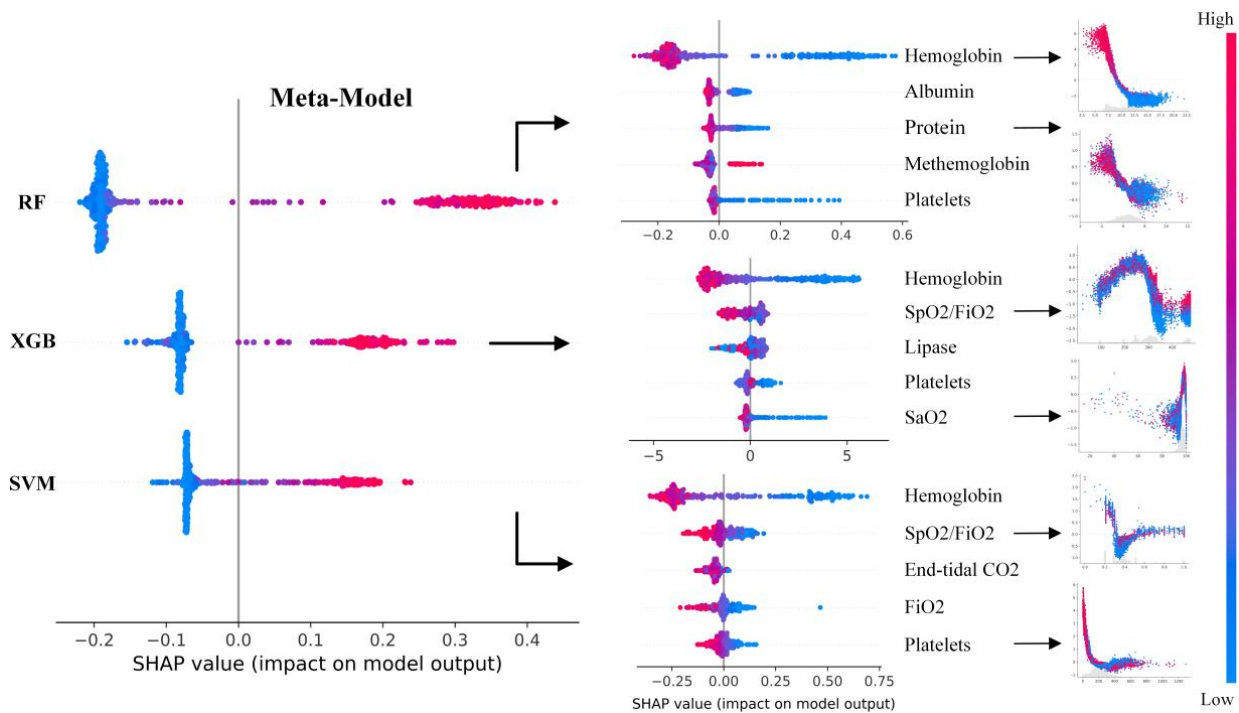

**Figure 4:** SHAP panel for the meta-model developed on the 2020 dataset.

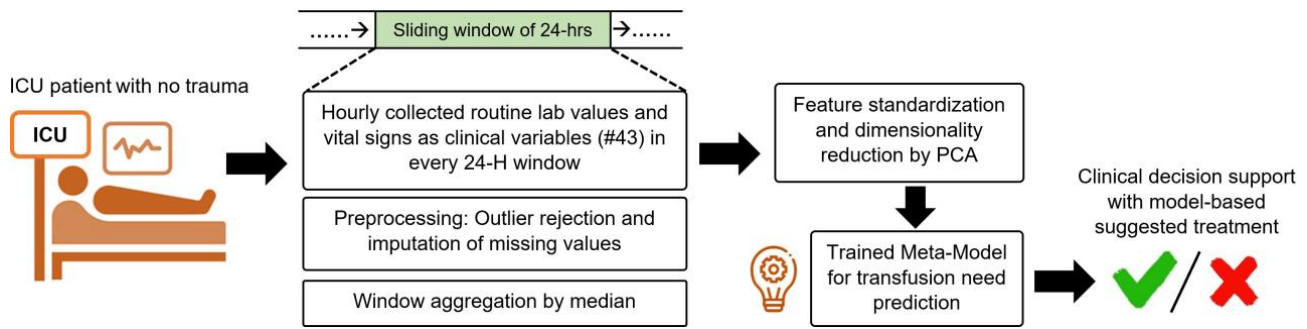

**Figure 5:** A use-case scenario of the developed meta-model includes collecting routine lab values and vital signs in a 24-hour sliding window. This data is then processed through a preprocessing workflow, preparing it as input for machine learning models. Overall, a complex, critically ill non-trauma patient poses a clinical decision for the need for blood transfusion. The primary objective is to employ a robust machine learning model, trained on historical real-world data, to predict the clinically relevant likelihood of requiring a blood transfusion. This prediction aims to aid clinicians in enhancing their decision-making process.

## A. Supplemental

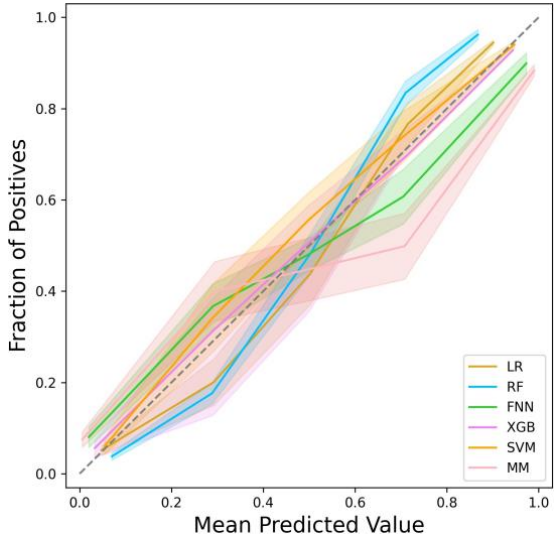

**Figure 1:** Calibration curves of the machine learning models using the five development scenarios.

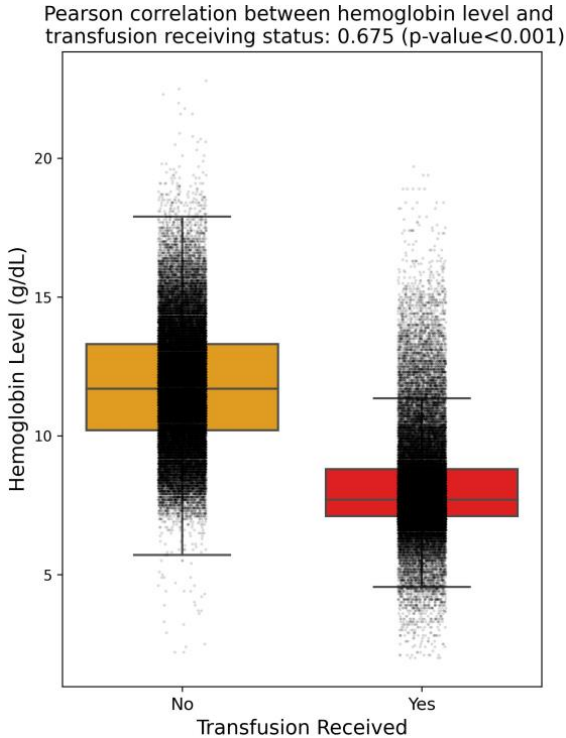

**Figure 2:** Distribution of hemoglobin level for transfused and non-transfused cohorts and cross-correlation of hemoglobin levels and blood transfusion decision. Each dot refers to the hemoglobin level of an ICU patient.

**Table 1**

List of routine clinical features selected in the study; variable names, their meanings, and measuring units.

| Variable names and their meanings | Unite |
| --- | --- |
| temperature: Body temperature | $^0$C |
| sbp_cuff: Cuff-based systolic blood pressure | mmHg |
| dbp_cuff: Cuff-based diastolic blood pressure | mmHg |
| pulse: Pulse rate (beats per minute) | beats per minute |
| unassisted_resp_rate: Respiratory rate | breaths per minute |
| spo2: Blood saturated oxygen concentration, SpO2 level | % |
| end_tidal_co2: End-tidal CO2 | mmHg |
| bicarb_(hco3): Bicarbonate | mmol/L |
| blood_urea_nitrogen_(bun) | mg/dL |
| chloride | mEq/L |
| creatinine | mg/dL |
| glucose | mmol/L |
| magnesium | mg/dL |
| osmolarity | mOsm/kg |
| phosphorus | mg/dL |
| potassium | mEq/L |
| sodium | mEq/L |
| hemoglobin | g/dL |
| met_hgb | g/dL |
| platelets | ×10$^9$/L |
| white_blood_cell_count | ×10$^9$/L |
| carboxy_hgb | % |
| alanine_aminotransferase_(alt) | U/L |
| albumin | g/L |
| alkaline_phosphatase | IU/L |
| bilirubin_direct | mg/dL |
| bilirubin_total | mg/dL |
| inr: International normalized ratio | - |
| lactic_acid | mmol/L |
| partial_prothrombin_time_(ptt) | s |
| protein | g/dL |
| lipase | U/L |
| b-type_natriuretic_peptide_(bnp) | pg/ml |
| troponin | ng/ml |
| fio2: Fraction of inspired oxygen | range: 0-1 |
| partial_pressure_of_carbon_dioxide_(paco2) | mmHg |
| partial_pressure_of_oxygen_(pao2) | mmHg |
| ph | - |
| saturation_of_oxygen_(sao2) | % |
| hemoglobin_a1c | % |
| best_map: Mean arterial pressure | mmHg |
| pf_sp: SpO2/FiO2 ratio | - |
| pf_pa: PaO2/FiO2 ratio | mmHg |

**Table 2**

Hyperparameter search space for tuning the models.

| Models | Hyperparameters | Search Space |
| --- | --- | --- |
| RF | Number of trees in the forest | {100, 150, 200, 300, 500, 1000, 1500, 3000} |
| | Minimum sample split | {2, 4, 5, 10} |
| | Maximum depth | {5, 8, 10, 12, 15, 20} |
| SVM | Kernel type | {linear, poly, sigmoid, rbf} |
| | Regularization parameter | {0.2, 0.5, 0.8, 1, 1.5, 3, 5, 10, 25, 50} |
| XGB | Learning rate | {0.01, 0.1} |
| | Number of boosting stages | {100, 250, 500} |
| | Maximum depth | {5, 7, 12, 15} |
| | Gamma | {0, 0.1, 1} |
| FNN | Number of hidden layers | {3, 4} |
| | Number of neurons | $\{z_n = 16 + 4(n - 1) \mid n \in \mathbb{Z}, 1 \le n \le 61\}$ |
| MM | Meta-model | {LR, RF, Adaboost, voting classifier, FNN, BN} |
| | Variance smoothing | {1e-9, 1e-7, 1e-9, 1e-5, 1e-3, 0.1, 0.5} |

Abbreviations used – FR: random forest, SVM: support vector machine, XGB: XGBoost, FNN: feedforward neural networks, MM: meta-model